\documentclass[letterpaper]{article} 
\usepackage{aaai25}  
\usepackage{times}  
\usepackage{helvet}  
\usepackage{courier}  
\usepackage[hyphens]{url}  
\usepackage{graphicx} 
\usepackage{natbib}  
\usepackage{caption} 
\usepackage{amsfonts,amssymb}
\usepackage{amsmath}
\usepackage{booktabs}
\usepackage{subcaption}
\usepackage{cleveref}

\usepackage{algorithm}
\usepackage{algorithmic}

\usepackage{newfloat}
\usepackage{listings}
\DeclareCaptionStyle{ruled}{labelfont=normalfont,labelsep=colon,strut=off} 
\floatstyle{ruled}
\newfloat{listing}{tb}{lst}{}
\floatname{listing}{Listing}
\title{MaFeRw: Query Rewriting with Multi-Aspect Feedbacks for Retrieval-Augmented Large Language Models}
\author{
   Yujing Wang\textsuperscript{\rm 1,\rm 2},  
    Hainan Zhang\textsuperscript{\rm 1,\rm 2}\thanks{\;Corresponding author.}, 
    Liang Pang\textsuperscript{\rm 4}, 
    Binghui Guo\textsuperscript{\rm 2},
    \\Hongwei Zheng\textsuperscript{\rm 3},
    Zhiming Zheng\textsuperscript{\rm 1,\rm2}
}
\affiliations{
    \textsuperscript{\rm 1}Beijing Advanced Innovation Center for Future Blockchain and Privacy Computing \\
    \textsuperscript{\rm 2} Institute of Artificial Intelligence, Beihang University, China\\
    \textsuperscript{\rm 3}Beijing Academy of Blockchain and Edge Computing, China\\
    \textsuperscript{\rm 4}Institute of Computing Technology, Chinese Academy of Sciences, Beijing, China \\
    \{wangyujing, zhanghainan\}@buaa.edu.cn
}

\usepackage{bibentry}

\begin{document}

\maketitle

\begin{abstract}
In a real-world RAG system, the current query often involves spoken ellipses and ambiguous references from dialogue contexts, necessitating query rewriting to better describe user's information needs. However, traditional context-based rewriting has minimal enhancement on downstream generation tasks due to the lengthy process from query rewriting to response generation. Some researchers try to utilize reinforcement learning with generation feedback to assist the rewriter, but this sparse rewards provide little guidance in most cases, leading to unstable training and generation results.
We find that user's needs are also reflected in the gold document, retrieved documents and ground truth. Therefore, by feeding back these multi-aspect dense rewards to query rewriting, more stable and satisfactory responses can be achieved. In this paper, we propose a novel query rewriting method MaFeRw, which improves RAG performance by integrating multi-aspect feedback from both the retrieval process and generated results. Specifically, we first use manual data to train a T5 model for the rewriter initialization. Next, we design three metrics as reinforcement learning feedback: the similarity between the rewritten query and the gold document, the ranking metrics, and ROUGE between the generation and the ground truth. Inspired by RLAIF, we train three kinds of reward models for the above metrics to achieve more efficient training. Finally, we combine the scores of these reward models as feedback, and use PPO algorithm to explore the optimal query rewriting strategy.
Experimental results on two conversational RAG datasets demonstrate that MaFeRw achieves superior generation metrics and more stable training compared to baselines.
\end{abstract}

%

\section{Introduction}
Retrieval-augmented generation (RAG) effectively addresses the issue of factual inaccuracy in large language models (LLMs) by integrating relevant retrieved information. It is widely used in various fields such as QA systems~\cite{rag_qa:2,rag_qa:3,rag_qa:1}, content generation~\cite{rag_gen:1,rag_gen:2,wang2023automatic}, and virtual agents~\cite{rag_agent:1,rag_agent:2,rag_agent:3}. 
Real-world RAG systems often rely on multi-turn dialogues, where queries include spoken ellipses and ambiguous references from dialogue contexts~\cite{a:10}, making it challenging for RAG systems to accurately understand user intent~\cite{convRAG}.
In light of this challenge, effective query rewriting is necessary to better describe user information needs and ensure the accuracy of retrieval and generation in RAG.

\begin{figure}[!t]
\centering
\includegraphics[width=0.9\columnwidth]{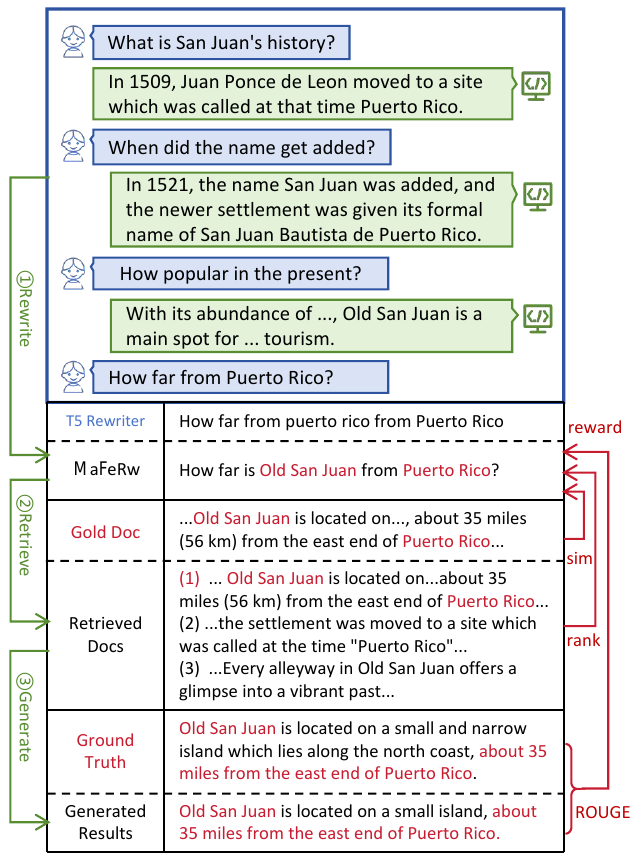} 
\caption{An example of MaFeRw serving RAG and the comparison with T5 rewriter. Green lines represent the inference process of rewriter and RAG, while Red lines indicate the three types of reward metrics feedback to MaFeRw.}
\label{intro}
\end{figure} 

Currently, RAG mainly performs context-based query rewriting to improve the quality of response generation, but the resulting enhancement is minimal due to the lengthy process from query rewriting to response generation. Although context-based query rewriting can directly affect the retrieval results, this impact is difficult to transmit to the final generated results through this loose-coupling retrieval and generation framework of RAG. As shown in Figure~\ref{intro}, the T5 rewriter~\cite{a:1} trained on context-based rewrite dataset fails to effectively capture the user's intent. To address this issue, \citeauthor{a:14} try to use reinforcement learning (RL) with generation feedback to assist the rewriter. They optimize the rewriter with policy gradient and use the accuracy and 'hit' rate of RAG inference's answers to form the reward. However, this reward is sparse and provides little guidance in most cases, resulting in unstable training and generation results of their model. Additionally, it is inefficient to directly use the full RAG inference to obtain RL feedback. Because RL typically requires a large number of training steps to achieve good results, and each inference in RAG takes a long time. Thus, if a stable and efficient method for training the rewriter can be designed, it will undoubtedly further enhance the effectiveness of RAG.

We find that user information needs are also reflected in the gold document, retrieved documents and ground truth, which allows for the design of multi-aspect rewards to provide more dense feedback signals. As illustrated in Figure~\ref{intro}, the current query 'How far from Puerto Rico' omits 'Old San Juan' from the previous context. The user intent 'Old San Juan's distance from Puerto Rico' is evident not only in the context but also in the gold document, retrieved documents, and ground truth. If query rewriting can leverage this information, it can better assist the model in generating correct answers. Therefore, by feeding multi-aspect dense rewards from the gold document, ranking measures, and generation metrics back to rewriting, more stable and satisfactory responses can be achieved.

In this paper, we propose MaFeRw, a novel query Rewriting method, to improve RAG performance by integrating Multi-aspect Feedback from both the retrieval process and generated results. Specifically, we first use manual data to train a T5 model~\cite{t5} as the rewriter's initialization. Next, we design three metrics as reinforcement learning feedback: the similarity between the rewritten query and the gold document, the ranking metrics of similarity between retrieved documents and the ground truth, and ROUGE between the generated response and the ground truth. Inspired by RLAIF~\cite{rlaif}, to achieve more efficient training, we collect datasets based on these metrics and train three kinds of reward models for them. Additionally, we use ROUGE scores of model-rewritten queries and manual-rewritten queries as the fourth feedback to measure the rewriter's performance. Finally, we combine scores of three reward models and the rewritten ROUGE as feedback, and use PPO algorithm \cite{ppo_1,ppo_2} to explore the optimal query rewriting strategy.

Experimental results on two conversational RAG datasets demonstrate that MaFeRw achieves superior generation metrics compared to baselines. Further analysis shows that multi-aspect dense rewards provide a more stable training process and generation results than single reward, validating the stability and transferability of MaFeRw.

The innovations of this paper are as follows:
\begin{itemize}
\item We find that dense reward feedback can lead to more stable and satisfactory generation results than single reward, and design four rewards from multiple aspects, such as gold document, retrieved documents and ground truth. 
\item To achieve more efficient RL training, we collect the feedback signals and train corresponding reward models, rather than using full RAG inference like the baseline.
\item Experimental results on two RAG datasets demonstrate that MaFeRw achieves superior generation metrics and more stable training than baselines.
\end{itemize}

\section{Related Work}
In dialogue systems, user utterances often contain omissions and ambiguous references~\cite{fang2022spoken,zhang2019recosa}. Therefore, a rewriting model is needed to resolve these ambiguities in the current query and recover missing elements (e.g., anaphora) from the context to better articulate the user's information needs. Current research on query rewriting mainly focuses on conversational search tasks, conversational QA tasks, and RAG tasks.

In conversational search tasks, researchers try to explore effective query rewriting methods to enhance the retrieval process. Some studies focus on extracting user intents from the dialogue context. For example, \citeauthor{a:1} use human rewrites as labels to train a sequence-to-sequence model as the rewriter. \citeauthor{a:9} propose query rewriting explicitly highlights relevant terms in the query context, facilitating contextual modeling and thereby improving the quality of the learned contextual query embeddings. In \cite{a:13}, LLM is used as the rewriter and is prompted to complete a "rewrite-then-edit" process based on the conversation history. Others incorporate retrieval feedback to enhance rewriting for retrieval tasks. \citeauthor{a:11} and \citeauthor{a:12} both take the feature similarity between the query and the target passage into the optimization objective when training the rewriting model and \citeauthor{a:8} use the retrieval effect as a reward for reinforcement learning. The rewritten queries are encouraged to achieve better retrieval performance.

In conversational QA tasks, mainstream query rewriting studies \cite{a:5,a:7} focus on feeding the generated results back into the rewriting model to improve the accuracy of the final answers. \citeauthor{a:5} train the QA model by regularizing the consistency of generated answers, so that the answers predicted by the model according to the original questions were similar to the answers generated according to the rewritten questions. \citeauthor{a:7} use the effect of generated answers as a reward for reinforcement learning, to improve the effect of rewriting on the final generated results.

In conversational RAG task, \citeauthor{a:14} firstly create a pseudo-dataset for supervised warm-up training, then further optimize the rewriter using a policy gradient RL framework, using the accuracy and 'hit' rate of LLMs' answers to form the reward.

However, for RAG tasks, focusing solely on the retrieval process results in rewrites unsuitable for prompting LLMs during generation. Conversely, considering only the feedback from the final generated results can lead to unstable training of the rewriting model due to the sparse reward signal. Therefore, we propose MaFeRw, which combines multi-aspect reward feedbacks to produce more stable and satisfactory responses.

\section{Problem Definition}
In RAG, each dialogue history $h^k$ contains a sequence of (query, answer) pairs $h^k=(q^1, a^1, ...,q^{k-1}, a^{k-1})$, where $q^i$ and $a^i$ denote the query and the system generation of the $i$-th turn. A conversational query $q^i$ can be elliptical and ambiguous. Given the dialogue history $h^k$ and the current query $q^k$, a query rewriting model $R_{\theta}$ with parameters $\theta$ rewrites $q^k$ to $q_R^{k}$:
$$q_R^{k}=R_{\theta}(h^k,q^k).$$
The rewritten query  $q_R^{k}$  is then fed into an off-the-shelf RAG system's retriever to search for relevant documents $D^k=\{d_1^{k},...,d_l^{k}\}$, where $d_i^{k}$ denotes the $i$-{th} retrieved document of the $k^{th}$ turn, $l$ is the number of retrieved documents. The current gold document is represented as $d_+^{k}$. Subsequently, the rewritten query $q_R^{k}$, the retrieved documents $D^k$, and the $(q^{k-1},a^{k-1})$ pair from the previous turn are combined as input to prompt the LLM to generate more satisfactory results.
The $(q^{k-1},a^{k-1})$ pair is used to instruct the LLM to mimic the style of the dataset by explicitly prompting the LLM to 'follow the style of the example responses in the context $(q^{k-1},a^{k-1})$'.
In this paper, we aim to train the query rewriting model $R_{\theta}$ to make the rewritten query  $q_R^{k}$ better describe user needs and ensure the accuracy of retrieval and generation in RAG.

\section{Approach}
\subsection{Framework Overview}
\begin{figure}[!t]
    \centering
    \begin{subfigure}{0.47\columnwidth}
    \centering   
      \includegraphics[width=1\linewidth]{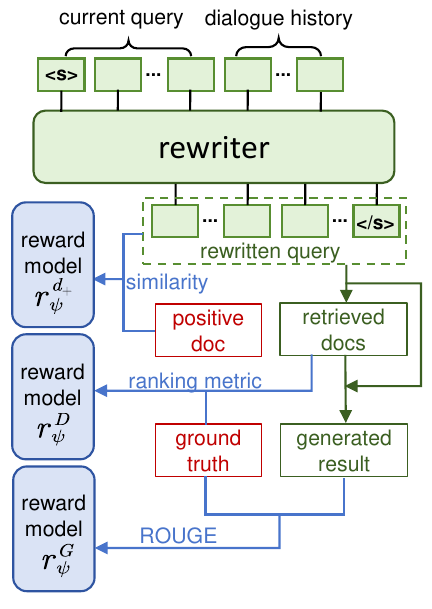}
        \caption{}
        \label{overview:rm}
    \end{subfigure}
    \begin{subfigure}{0.47\columnwidth}
    \centering   
      \includegraphics[width=1\linewidth]{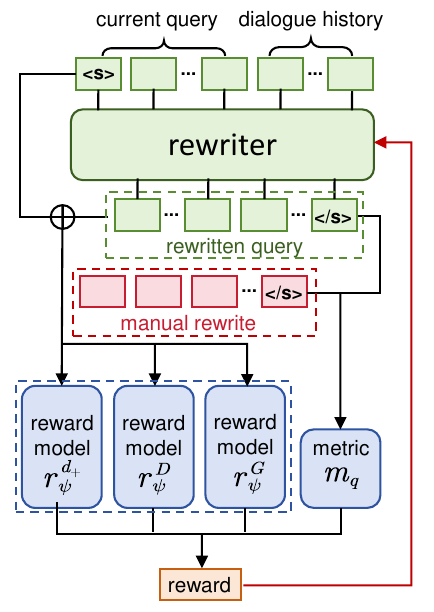}
        \caption{}
        \label{overview:ppo}
    \end{subfigure}
    \caption{The framework of MaFeRw. (a) Three feedback metrics are: the similarity between rewritten query and gold document, the ranking metric of similarity between ground truth and retrieved documents, and the ROUGE scores between generation and ground truth. Corresponding reward models are trained for these metrics. (b) When training the rewriter using PPO algorithm, the reward is composed of scores from three reward models and the rewritten ROUGE.}
    \label{fig:overview}
\end{figure}
For MaFeRw, the rewriter is trained to transform the current query $q^k$ into $q_R^{k}$ utilizing the dialogue history $h^k$, which enables the RAG system to generate more satisfactory results. Firstly, we use manual rewriting data to train a T5 model as the rewriter’s initialization. Secondly, we design three metrics as RL feedback: the similarity between the rewritten query $q_R^{k}$ and the gold document $d_+^k$, the ranking metrics of similarity between retrieved documents $D^k$ and the ground truth $G^k$, and ROUGE between the generation $g^k$ and the ground truth $G^k$. Inspired by RLAIF, we train three kinds of reward models for the above metrics on three corresponding paired data sets, as shown in Figure \ref{overview:rm}. Considering the obvious improvement of manual rewriting in the generation, ROUGE between manual rewrites and model rewrites is used as another feedback. Finally, we combine scores of these reward models and the rewritten ROUGE to feedback PPO training, exploring the optimal query rewriting strategy, as shown in Figure \ref{overview:ppo}. Source code: https://github.com/yjEugenia/MaFeRw.

\subsection{Rewriter Initialization}
To equip the rewriter with fundamental rewriting capabilities, we train the pre-trained T5-base model on datasets containing manually rewritten queries. The query rewriter concatenates the current query $q^k$ with the dialogue history $h^k$ as input. Similar to the approach used by \cite{a:8}, a separator token "[SEP]" is inserted between each interaction, and dialogue turns are connected in reverse order, as $$I=\mathrm{[CLS]}q^k\mathrm{[SEP]}a^{k-1}\mathrm{[SEP]}q^{k-1}\cdots q^1\mathrm{[SEP]}.$$

When $q_+$ is a manual-rewritten query, the objection of initialization is to optimize parameters $\theta$ of the rewriter $R_{\theta}$ by minimizing the cross-entropy loss between the model's prediction $q_R$ and $q_+$: 
$$\mathcal{L}_{init}=-y_{q_+}\mathrm{log}y_{q_R},$$
where $y_{q_+}$ is the one-hot vector of $q_+$ and $y_{q_R}$ is the distribution over tokens in $q_R$ predicted by the rewriter. The rewriter after initialization is represented as $R_{\theta}^{0}$, which defines a probabilistic policy for this task.
\subsection{Multi-aspect Feedbacks}
Since user information needs are also reflected in gold documents, retrieved documents, and ground truth, we design three metrics as RL feedback. For simplicity, we omit the superscript $k$ of $q^k$, $q_R^{k}$, $d_+^k$, $D^k$, $G^k$ and $g^k$ in the rest of the paper if not specified.

\subsubsection{(1) The similarity between $q_R$ and $d_+$}
In RAG, the retriever converts the query and documents into dense vectors, calculates similarity scores between the query vector and document vectors, and retrieves the top-$l$ most relevant documents. To ensure the rewritten query $q_R$ retrieves the gold document $d_+$ more effectively, the cosine similarity (CS) between the dense vector $v_{q_R}$ of $q_R$ and the dense vector $v_{d_+}$ of $d_+$ is used as a metric $m_{d_+}$ which provides retrieval feedback to the rewriter, i.e., $m_{d_+}=\mathrm{CS}(v_{q_R},v_{d_+})$.
\subsubsection{(2) The ranking metric of similarity between $D$ and $G$}
Since user needs are embedded in the ground truth $G$, and the order of retrieved documents $D$ impacts generation quality, we hope the rewritten query can assist the retrieval model in prioritizing the document that exhibits high semantic relevance to the ground truth. Therefore, we propose a ranking metric $m_{D}$ using $G$ and $D=\{d_1,\cdots,d_l\}$. 
Specifically, we transform the ground truth $G$ into a dense vector $v_G$ using the retrieval embedding model. Then the CS value is calculated between $v_G$ and dense vectors of retrieved documents $\{v_{d_i}\}_{i=1}^l$. For each $d_i$, let $1/i$ be its rank score. The metric $m_{D}$ is computed by summing the product of each document's rank score and its CS value with $v_G$, formulated as $m_{D} =\Sigma_{i=1}^l 1/i \cdot\mathrm{CS}(v_G,v_{d_i})$. A higher $m_{D}$ indicates that the $d_i$ with greater similarity to $G$ is ranked higher, thereby facilitating the generation of better responses.

\subsubsection{(3) ROUGE between $g$ and $G$}
To ensure that the generated output $g$ produced by RAG closely aligns with the ground truth $G$, we use ROUGE score between the generation $g$ and the ground truth $G$ as the metric $m_{G}$.
\subsubsection{(4) ROUGE between $q_R$ and $q_+$}
Considering the obvious improvement of manual rewrites in the generation, we also use ROUGE between the manual rewrite $q_+$ and model rewrite $q_R$ as a reference metric $m_q$. 
\subsection{Reward Model}
The end-to-end RL approach without a reward model demands extensive training time due to the prolonged inference time of RAG and the large number of training steps required. Incorporating the complete RAG process into RL led to training durations of several days, severely limiting tuning and flexibility in later stages. To address this issue, reward models were employed to evaluate rewritten queries, effectively decoupling the evaluation from RAG process and enabling more efficient training.
Inspired by RLAIF, we train three types of reward models for the aforementioned metrics on three corresponding paired datasets. The reward model is based on the T5-base model with a value head, possessing significantly fewer parameters than the generation model Llama2-13b. During RL training, the reward is derived from scores of reward models rather than using the complete RAG process, thereby reducing RAG inference time.

Specifically, for the given dialogue history $h$ and the current query $q$, we pair the corresponding manual rewrite $q_{m}$, rewrite $q_{T5}$ generated by the initialized rewriter, and rewrite $q_s$ obtained through sampling. All rewrites are then put into the RAG system to calculate values of $m_{d_+}$, $m_{D}$ and $m_{G}$. For each metric, we compare rewrites in every pair and label the one with the higher metric value as 'chosen' and the other as 'rejected', creating a paired dataset of (chosen, rejected) rewrites. Taking the paired dataset of metric $m_{G}$ as an example, if $m_{G}(q_{m})>m_{G}(q_{T5})>m_{G}(q_{s})$, then for the dialogue history $h$ and current query $q$, three pairs of data are collected:
\begin{align}
\{&(chosen: [h+q, q_{m}], rejected: [h+q, q_{T5}]), \nonumber \\
& (chosen: [h+q, q_{m}], rejected: [h+q, q_{s}]),\nonumber \\
&(chosen: [h+q, q_{T5}], rejected:[h+q, q_{s}])\}, \nonumber
\end{align}
here "+" denotes string concatenation.
When training the reward model, we adopt a method similar to \citeauthor{ppo_rm}. By following the Bradley-Terry model \cite{ppo_rm_rule}, we formulate a preference distribution by employing the reward model $r_{\psi}$ as outlined below:
\begin{equation}
    P_{\psi}(q_c\succ q_r|h+q)=\sigma(r_{\psi}(h+q,q_c)-r_{\psi}(h+q,q_r)),
\end{equation}
where $\sigma$ is the logistic function, $q_c$ and $q_r$ represent the chosen and rejected rewrite, respectively. This problem can be treated as a binary classification task, resulting in a negative log-likelihood loss function:
\begin{equation}
    \mathcal{L}_{rm}=-\mathbb{E}_{D_{rm}}[\mathrm{log}\sigma(r_{\psi}(h+q,q_c)-r_{\psi}(h+q,q_r))],
\end{equation}
where $D_{rm}$ is the paired dataset. 

In this work, we use the initialized rewriter to initialize $r_{\psi}$. Additionally, we incorporate an extra linear layer on top of the final transformer layer to generate a scalar prediction representing the reward value. Let $r_\psi^{d_+}$, $r_\psi^{D}$ and $r_\psi^{G}$ denote the reward models associated with metrics $m_{d_+}$, $m_{D}$ and $m_{G}$, respectively. Given the dialogue history $h$, the current query $q$, and the rewritten query $q_R$, the corresponding rewards can be obtained as $r_\psi^{d_+}(h+q, q_R)$, $r_\psi^{D}(h+q, q_R)$ and $r_\psi^{G}(h+q, q_R)$, abbreviated as $r_{d_+}(q_R)$, $r_{D}(q_R)$ and $r_{G}(q_R)$. To facilitate subsequent aggregation, we scale the scores of reward models to [0, 1].
\subsection{RL Training}
To further train the rewriter, we employ a policy gradient RL framework. In the context of RL, the optimization of the rewriter can be viewed as a Markov Decision Process (MDP) represented by the 5-tuple $\langle S, A, P, r, \gamma \rangle$. Here, the state space $S$ is a finite set constrained by the vocabulary and the sequence length. The action space $A$ comprises the available vocabulary and the transition probability $P$ is determined by the policy network, specifically the rewriter model $R_{\theta}^{RL}$. The reward function $r$ provides the reward value based on the current state, and $\gamma$ is the discount factor.
For each step $t$, the current state $s^t$ comprises the dialogue history $h$, the current query $q$, and the already generated rewrite $q_{R,[<t]}$. The action $a^t$ is the generation of the next token based on the current state. When the generation process stops, the reward for this episode is calculated using the reward function.

In the RL stage, for the dialogue history $h$, the current query $q$, and the rewritten query $q_R$, we combine scores of these reward models and the metric $m_q$ as the feedback $r_{RL}$ for RL training. This is expressed as:
\begin{align}
r_{RL}(q_R)=\lambda_1 r_{d_+}(q_R) + \lambda_2 r_{D}(q_R)
\\+ \lambda_3 r_{G}(q_R)+ \mu m_q(q_R), \nonumber
\end{align}
where $\lambda_1$, $\lambda_2$ and $\lambda_3$ are hyperparameters that control the relative importance of each reward model, and $\mu$ is a hyperparameter that controls the weight of the metric $m_q$.

Then we use PPO algorithm to explore the optimal query rewriting strategy $R_{\theta}^{RL}$. The reward objective we need to maximize in policy optimization is:
\begin{equation}
    r_{total}=r_{RL}(q_R)-\eta\mathrm{KL}(R_{\theta}^{RL}||R_{\theta}^0),
\end{equation}
where $\eta$ is a coefficient governing the magnitude of KL penalty, the initial policy model is the initialized rewriter.

\section{Experiments}
\subsection{Experimental Settings}
\subsubsection{Datasets}
We conduct main experiments on two multi-turn dialogue RAG datasets, including QReCC \cite{qrecc} and TopiOCQA \cite{topiocqa}. And conduct the transferability experiment on the WSDM@24 Multi-Doc QA dataset \footnote{https://sites.google.com/view/wsdm24-docqa}. QReCC contains 14k conversations with 81k question-answer pairs, built on questions from TREC CAsT \cite{trec}, QuAC \cite{quac}, and Google Natural Questions \cite{nq}. TopiOCQA is an open-domain conversational dataset with topic switches based on Wikipedia, containing 3,920 conversations with information-seeking questions and free-form answers.  WSDM@24 Multi-Doc QA dataset comprises multi-turn question-answering dialogues and corresponding documents from Xiaohongshu. 
\subsubsection{Evaluation Metrics}
To evaluate the generated results of RAG, we use four standard evaluation metrics: ROUGE-1, ROUGE-L \cite{rouge}, BLEU \cite{bleu}, and METEOR \cite{meteor}. ROUGE prioritizes recall and n-gram overlap, BLEU balances precision and brevity, and METEOR emphasizes semantic similarity. In addition, we utilize MRR \cite{mrr} to evaluate the retrieval results, as it measures the system's effectiveness in placing relevant items at the top of the ranked list. 
\subsubsection{Baselines}
We compare MaFeRw with four conversational RAG methods: raw RAG \cite{rawrag}, T5-base rewriter \cite{a:1}, RL-base rewriter \cite{a:14}, and ConvGQR \cite{a:11}. Raw RAG is the baseline model that performs retrieval and response generation without query rewriting, using the concatenation of dialogue history and the current query. T5-base rewriter uses manual rewrites to train a T5 model for query rewriting based on dialogue history. RL-base rewriter utilizes RL with generation feedback to assist the training of the rewriter. ConvGQR generates potential answers for query expansion and incorporates similarity between the rewrite and gold document into the rewriter's optimization objective.
\subsubsection{Implementation Details}
In this work, we train the query rewriter based on the pre-trained T5-base model. The dense retrieval component in the RAG system is constructed using FAISS \cite{faiss} and the pre-trained embedding model msmarco-roberta-base-ance-firstp \cite{embedding_model}. The generative model is the Llama-2-13b-chat model \cite{llama2}. After initialization of the rewriter, a value head is added to the rewriter to initialize reward models. 
Details on hyperparameter determination are provided in the Appendix\footnote{Appendix can be found at: https://arxiv.org/abs/2408.17072}.
\subsection{Main Results}
We demonstrate experimental results on two datasets. 
\subsubsection{Accuracy of Reward Models}
\begin{table}[t]
\footnotesize 
\centering
\begin{tabular}{cccc}
\toprule
    & $r_{d_+}$ & $r_{D}$ & $r_{G}$ \\
\midrule
QReCC&  0.8927& 0.8534& 0.7802\\
TopiOCQA& 0.8362& 0.7303& 0.7277\\
\bottomrule
\end{tabular}
\caption{The accuracy of reward models.}
\label{rm-table}
\end{table}
It is crucial to train reward models to accurately score a given rewritten query with dialogue history and correctly categorize it as either "chosen" or "rejected". 
Table \ref{rm-table} gives information about the accuracy of trained metric-specific reward models on their respective test sets. On both QReCC and TopiOCQA datasets, classification accuracies consistently exceeded 70\% for all reward models, ensuring the success of subsequent PPO training based on these reward models.

\begin{table}[t]
\footnotesize 
\setlength{\tabcolsep}{1mm}
\centering
\begin{subtable}[t]{\linewidth}
\centering
\begin{tabular}{lccccc}
\toprule
   Method & ROUGE-1 & ROUGE-L & BLEU & METEOR & MRR \\
\midrule
raw RAG& 31.98& 26.44& 5.776& 35.10 & 0.4729\\
T5-base & 35.73& 30.50& 11.07& 38.74 & 0.4884\\
RL-base & 35.47& 29.92& 11.27& 38.93 & 0.4742\\
ConvGQR & 34.41& 28.33& 10.39& 37.53 & 0.4974\\
MaFeRw& \textbf{37.05}& \textbf{31.31}& \textbf{12.20}& \textbf{40.76} & \textbf{0.5032}\\
\midrule
Manual& 39.73& 34.07& 13.54&  43.85 & 0.5649\\
\bottomrule
\end{tabular}
\subcaption{QReCC}
\end{subtable}
\vspace{3mm}

\begin{subtable}{\linewidth}
\centering
\begin{tabular}{lccccc}
\toprule
   Method & ROUGE-1 & ROUGE-L & BLEU & METEOR & MRR \\
\midrule
raw RAG& 20.76& 17.70& 2.295&  25.48 &  0.2664\\
T5-base &  21.46& 19.12&3.854&  25.92& 0.3015\\
RL-base &  22.94& 20.46&4.993& 27.83& 0.3717\\
ConvGQR & 17.05& 15.11& 3.503& 22.16 & 0.3081\\
MaFeRw& \textbf{23.97}& \textbf{21.38}& \textbf{5.496}& \textbf{29.51} & \textbf{0.3802}\\
\midrule
Manual& 24.72& 21.95& 5.367&  30.00 &  0.4001\\
\bottomrule
\end{tabular}
\subcaption{TopiOCQA}
\end{subtable}
\caption{The comparison of MaFeRw and the baselines on QReCC and TopiOCQA datasets.}
\label{result-table}
\end{table}
\subsubsection{RAG Performance}
Table \ref{result-table} presents a comparison of our method and the baseline approaches in terms of retrieval and generation performance. Compared to the baselines, MaFeRw demonstrates significantly improved retrieval and generation performance on both datasets. MaFeRw achieves 5.07 and 3.19 ROUGE-1 higher than raw RAG on QReCC and TopiOCQA, respectively. Moreover, in both datasets, the improvement in MRR score is accompanied by an improvement in generation metrics. 

Compared to QReCC dataset, TopiOCQA dataset features topic shifts within the dialogues. Notably, on TopiOCQA, the rewritten queries show a more significant improvement in MRR compared to using raw RAG with the dialogue history directly for retrieval. MRR of Long Rewriter improves by 6.41\% on QReCC compared to raw RAG, and by 42.66\% on TopiOCQA. Even T5-base rewriter achieves a more pronounced MRR improvement on TopiOCQA compared to QReCC. However, due to the topic shifts in TopiOCQA, while ConvGQR’s query expansion based on generated answers improves retrieval quality, it fails to effectively enhance generation quality. These findings suggest that query rewriting can effectively filter out redundant information, thereby significantly enhancing retrieval performance when dealing with dialogues involving topic shifts.

The dialogues in QReCC come from various sources, resulting in more diverse question-and-answer styles than those in TopiOCQA. As a result, RL-base rewriter struggles to adapt to these stylistic variations in QReCC dataset, leading to lower performance metrics compared to T5-base rewriter, and revealing its instability in training and outcomes. These results verify that more satisfactory responses can be obtained by feeding back the results of the remote retrieval and generation modules to query rewriting.

Additionally, it can be observed that the RAG performance with manual rewrites remains consistently strong across these two datasets. This observation underlies our decision to incorporate ROUGE between the rewriter output and the manual rewrite in the RL feedback.
\subsection{Ablation Study}

\begin{table}[t]
\footnotesize
\setlength{\tabcolsep}{1mm}
\centering
\begin{subtable}{\linewidth}
\centering
\begin{tabular}{ccccc}
\toprule
    & ROUGE-1 & ROUGE-2& ROUGE-L & MRR \\
\midrule
$m_{d_+}$ & 36.10& 20.70& 30.22& 0.4863\\
$m_{D}$ & 35.69& 20.40& 29.96& 0.4924\\
$m_{G}$& 36.69& 21.59& 31.03&  0.4937\\
$m_{q}$ & 36.88& 22.06& 31.22& 0.4897\\
MaFeRw&37.04&21.84&31.31&0.5032\\
\midrule
T5-base rewriter & 35.73& 20.02& 30.50&  0.4884\\
\bottomrule
\end{tabular}
\subcaption{QReCC}
\end{subtable}
\vspace{3mm}

\begin{subtable}{\linewidth}
\centering
\begin{tabular}{ccccc}
\toprule
    & ROUGE-1 & ROUGE-2& ROUGE-L & MRR \\
\midrule
$m_{d_+}$ & 21.27& 10.24& 18.78& 0.3619\\
$m_{D}$ & 21.86& 10.41& 19.24& 0.3693\\
$m_{G}$& 23.14& 11.21& 20.55&  0.3570\\
$m_{q}$ & 22.92& 10.96& 20.23& 0.3723\\
MaFeRw&23.97&12.31&21.38&0.3802\\
\midrule
T5-base rewriter & 21.46& 9.874& 19.12&  0.3015\\
\bottomrule
\end{tabular}
\subcaption{TopiOCQA}
\end{subtable}
\caption{The impact of four different metrics on the RL training of the query rewriter.}
\label{ablation-table}
\end{table}

We conduct RL training on both datasets using feedback from a single reward model or metric and compare results with those of T5-base rewriter. Table \ref{ablation-table} demonstrates the impact of four different metrics on rewriter's RL training. 
\subsubsection{(1) Impact of the metric $m_{d_+}$}
For QReCC, we can be see that when only the reward model $r_{d_+}$ is used, ROUGE-1 and ROUGE-2 scores are higher than those of T5-base rewriter but MRR is lower. For TopiOCQA, MRR is much higher than that of T5-base rewriter. This indicates that using $m_{d_+}$ as feedback can, to some extent, guide RL optimization to achieve better retrieval and encourage LLM to improve generation quality, although this promoting effect is not stable. 
\subsubsection{(2) Impact of the metric $m_{D}$}
Using only the reward model $r_{D}$ yields a better MRR than T5-base rewriter but lower ROUGEs on QReCC. For TopiOCQA, both ROUGE and MRR of $r_D$ are higher than those of T5-base rewriter. This confirms that $m_{D}$ helps the retriever more effectively find relevant documents, but this benefit does not translate to the generative model, thus failing to produce responses that better meet user needs.
\subsubsection{(3) Impact of the metric $m_{G}$}
For both datasets, when only the reward model $r_{G}$ is used, ROUGE and MRR scores improve compared to T5-base rewriter. On TopiOCQA, ROUGEs show a clear advantage, even though MRR of $r_{G}$ is smaller than those of $r_{d_+}$ and $r_{D}$. This confirms the positive impact of $m_{G}$ on RAG quality and highlights the necessity of conjunction with other metrics to achieve better results.
\subsubsection{(4) Impact of the metric $m_{q}$}
When only the metric $m_{q}$ is used, both ROUGEs and MRR scores show improvement over T5-base rewriter. On QReCC dataset, the improvement in ROUGE-1 score is notable. This suggests that when facing multi-source, diverse-style conversational data like QReCC, the metric $m_{q}$ can guide RL optimization to achieve better RAG performance.

\subsubsection{}These results indicate that individual metrics can improve certain aspects of RAG performance, but their performance varies across different datasets, which is insufficient to stably support RL training. Therefore, combining multiple metrics is necessary to achieve better overall results.

\subsection{Analysis}
\begin{figure}[!t]
    \centering
    \begin{subfigure}{0.49\columnwidth}
    \centering   
      \includegraphics[width=1\linewidth]{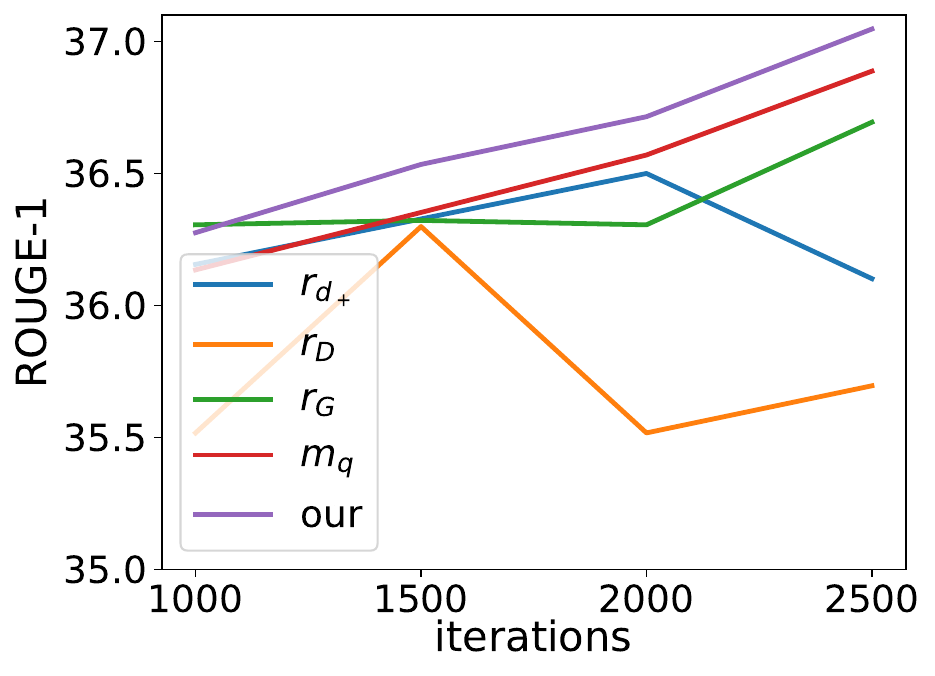}
        \caption{ROUGE-1}
        \label{fig:rouge}
    \end{subfigure}
    \begin{subfigure}{0.49\columnwidth}
    \centering   
      \includegraphics[width=1\linewidth]{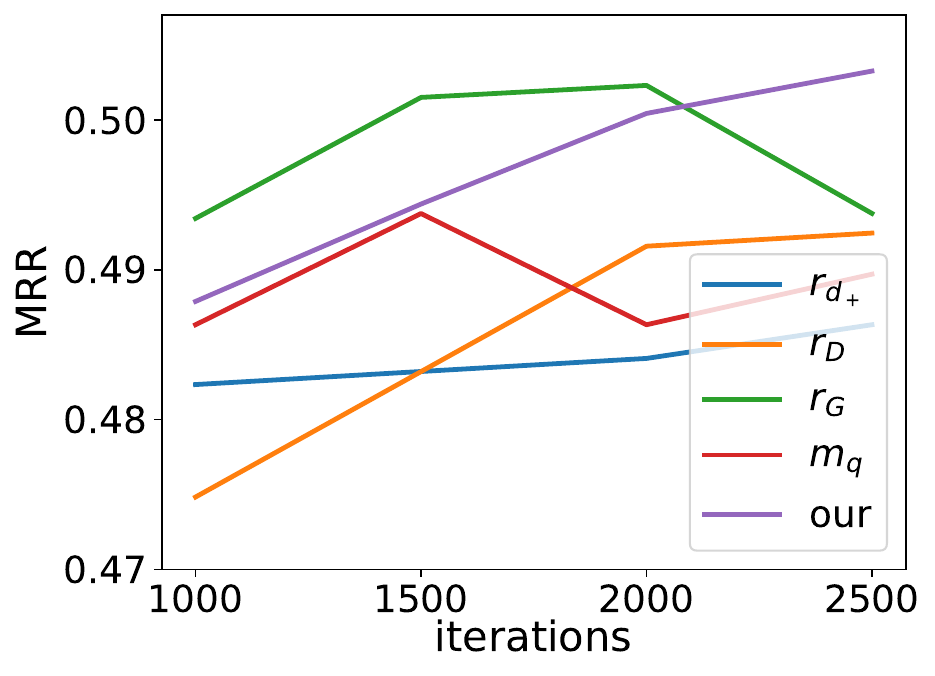}
        \caption{MRR}
        \label{fig:mrr}
    \end{subfigure}
    \caption{The changes on ROUGE-1 and MRR as training iterations increase when rewrites are applied to RAG.}
    \label{fig:analysis}
    
\end{figure}

\begin{figure}[!t]
\centering
\includegraphics[width=\columnwidth]{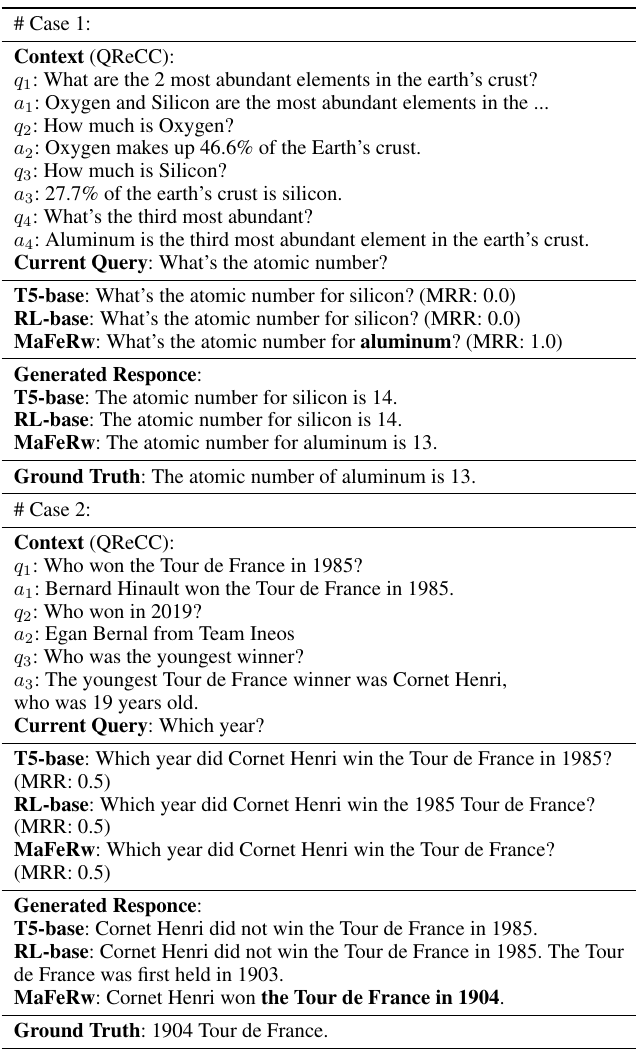} 
\caption{The case study comparing with two baselines.}
\label{case-table}
\end{figure}
To investigate whether using scores output by reward models and $m_{q}$ as RL feedback can ensure stable training results, we conduct experiments on QReCC. We use either a single metric or reward model to guide the rewriter's RL training and test the retrieval and generation performance of rewrites after 1000, 1500, 2000, and 2500 iterations. Figures \ref{fig:rouge} and \ref{fig:mrr} show the changes in ROUGE-1 and MRR scores as the number of iterations increases, respectively.

It can be observed that when training the rewriter using feedback from the reward model $r_{d_+}$ or $r_{D}$, MRR of the rewritten queries in RAG gradually increases, while ROUGE-1 shows slight decreases or fluctuations. This indicates that these two types of reward models can reliably guide the rewriter to optimize towards better retrieval performance, but are insufficient to consistently constrain the rewriter to improve generation results. 
When utilizing the reward model $r_{G}$ for training feedback, although there are fluctuations in ROUGE-1 and MRR metrics, there is an overall upward trend. This trend validates that the reward model can provide positive guidance to the rewriter. 
When directly using $m_{q}$ for feedback, ROUGE-1 exhibits a stable increase while MRR shows some variability. This indicates that the metric can, to some extent, consistently guide the rewriter towards optimizing in a manner beneficial for RAG generation. 
Furthermore, Figure \ref{fig:analysis} shows the changes in ROUGE-1 and MRR as iterations increase after training the rewriter with synthetic feedback. Both metrics show a more stable upward trend, further validating MaFeRw's stability. 

Based on the above results, it can be analyzed that, by training on paired datasets, reward models can capture a more essential correspondence between rewriting based on dialogue history and each metric. Therefore, RL training with the feedback given by reward models can obtain more stable and effective results than directly training with the whole process of RAG. 

\subsection{Case Study}
We present a case study and a comparison with two other rewrite baselines in Figure \ref{case-table} to illustrate more intuitively the improvement of MaFeRw in capturing users' needs for multi-turn dialogue RAG.

For the first case, according to the dialogue history, the real need of the current query is to ask for 'the atomic number of aluminum'. However, because ' silicon' is also mentioned in the dialogue history, the two baselines are disturbed and add 'silicon' mistakenly. As a result, the error retrieval and generation are triggered. 
In the second case, both baselines incorrectly add 'in 1985' to rewritten queries. Although the correct document can still be retrieved, the LLM is unable to give a correct reply based on the rewritten query due to the interference of the information '1985'. 

These results verify that MaFeRw can accurately capture the anaphora relationship between the current query and the dialogue history, especially when other interfering words are mentioned in the dialogue history.

\subsection{Transferability}
\begin{table}[t]
\footnotesize
\setlength{\tabcolsep}{1mm}
\centering
\begin{tabular}{lcccc}
\toprule
   Method & ROUGE-1 & ROUGE-L & BLEU & METEOR \\
\midrule
raw RAG& 38.79&  23.95& 5.989&  24.35\\
T5-base &  39.80& 24.74&7.227&  25.61\\
MaFeRw& \textbf{41.01}& \textbf{25.86}& \textbf{8.226}& \textbf{26.46}\\
\bottomrule
\end{tabular}
\caption{The generation performance on the WSDM@24 Multi-Doc QA task.}
\label{wsdm-table}
\end{table}
To evaluate the transferability of MaFeRw, we test its performance on the WSDM@24 Multi-Doc QA dataset by using the rewriting model trained on QReCC to rewrite the current query, as shown in Table \ref{wsdm-table}. This allows us to assess its effectiveness in handling multi-document conversational QA tasks.
It can be observed that compared to the T5-base rewriter, MaFeRw achieves better generation performance on this task. This validates that the rewriter trained using our method possesses generalization capability across different conversational tasks.

\section{Conclusion}
This paper introduces MaFeRw, a novel query rewriting method that enhances RAG performance by incorporating multi-aspect feedbacks. We find that dense reward feedback can lead to more stable and satisfactory generation results than single reward, and design four rewards from multiple aspects, such as gold document, retrieved documents and ground truth. The scores of these reward models are combined to optimize the rewriter using PPO algorithm.
Experimental results on two conversational RAG datasets show MaFeRw outperforms baselines in generation metrics, and further analysis validates MaFeRw's stability and transferability.
Future work involves document re-ranking to better match retrieved documents with contextual constraints and prompt reconstruction to capture users' complex intents in multi-turn dialogues. 

\section*{Acknowledgments}
This work was funded by the National Natural Science Foundation of China (NSFC) under Grants No. 62406013, the  Beijing Advanced Innovation Center Funds for Future Blockchain and Privacy Computing(GJJ-23-006) and the Fundamental Research Funds for the Central Universities. 
\appendix
\section{Details of Experimental Settings}
The experiments are conducted on one Nvidia A100 40G GPU. We implement all models by PyTorch \cite{PyTorch}, LangChain \cite{LangChain} and Huggingface's Transformers \cite{huggingface}. 
For rewriter initialization, we employ the Adam optimizer \cite{adam} and train for 10k iterations with a learning rate of 5e-6 and a batch size of 16. Then we add a value head to the initialized rewriter to initialize reward models. Reward models are trained for 5 epochs using the Adam optimizer and the cosine-type scheduler. The learning rate is set to 5e-5 and the batch size is 16. During the RL phase, the learning rate is set to 1.41e-5, the batch size is 32, and the model is trained for 80k steps, with the parameter $\eta$ set to 0.05. For QReCC dataset, the hyperparameters $\lambda_1$, $\lambda_2$, $\lambda_3$, and $\mu$ are set to 0.04, 0.01, 0.35, and 0.6. For TopiOCQA dataset, the values are 0.04, 0.01, 0.45, and 0.5, respectively. 

For the selection of hyperparameters in the reward function, we first refer to the improvement of retrieval and generation performance when a certain metric is used alone, and make the initial assignment according to the proportion, then use grid search to fine-tune and select the optimal parameters based on experimental results. To better compare with other baselines like ConvGQR, the maximum generation length is set to 32. QReCC focuses on the query rewriting problem within conversational scenarios by approaching the human-rewritten query. Thus, it provides an oracle query for each conversation turn. TopiOCQA focuses on the challenge of the topic switch under conversational settings, whose sessions are longer than QReCC and thus present more difficulties for query reformulation. Different from QReCC, it does not provide human-rewritten queries. The publisher of TopiOCQA finetunes the T5 model on a rewrite of QReCC and uses it to generate a rewrite of TopiOCQA. Since this is obtained based on manual rewriting training, we take it as the label of TopiOCQA representing manual rewriting. During the retrieval process, the number of retrieved articles was set to 5. 

When training the reward model, we randomly selected 48\% of the whole training set as the reward model training set and 12\% as the test set instead of using all the data for the training of the reward model. This can better verify whether the reward model is overfitting. Due to hardware limitations, we filtered the document collections of QReCC and TopiOCQA datasets, retaining only the positive documents from the training and test sets, along with randomly sampled negative documents, while also removing any corrupted files.

We implement baselines based on our experimental setting and their open-source code and material. For the normal evaluation, we train T5-base, RL-base, and ConvGQR on the corresponding datasets rather than using external resources.
\bibliography{aaai25}

\end{document}